\documentclass[10pt, conference]{IEEEtran}

\usepackage{amsmath,amssymb,amsfonts}
\usepackage{graphicx}
\usepackage{pifont}

\usepackage{amsthm}
\usepackage{xcolor}
\usepackage[figuresright]{rotating}

\usepackage{graphicx}
\graphicspath{{/}{fig/}}

\usepackage{array}
\usepackage{textcomp}
\usepackage{xcolor}
\usepackage{multirow}
\usepackage{booktabs}

\usepackage{mathtools}
\usepackage{breqn}
\usepackage{float}

\usepackage{tikz,pgfplots}
\pgfplotsset{compat=1.7}
\usepgfplotslibrary{groupplots}

\usepackage{caption}
\usepackage{subcaption}

\usepackage{multirow,tabularx}
\usepackage{hyperref}

\usepackage{algorithmic}
\usepackage[vlined, ruled, shortend]{algorithm2e}

\usepackage[capitalise]{cleveref}

\newlength\figureheight
\newlength\figurewidth
\SetAlCapNameFnt{\footnotesize}
\SetAlCapFnt{\footnotesize}

\title{
    Towards Lifelong Federated Learning in Autonomous Mobile Robots with Continuous Sim-to-Real Transfer
}

\author{
    \IEEEauthorblockN{
        \vspace{1em}
        Xianjia Yu\IEEEauthorrefmark{1},
        Jorge Pe\~na Queralta\IEEEauthorrefmark{1},
        Tomi Westerlund\IEEEauthorrefmark{1}
    }
    \IEEEauthorblockA{
        \normalsize
        \IEEEauthorrefmark{1}\href{https://tiers.utu.fi}{Turku Intelligent Embedded and Robotic Systems (TIERS) Lab, University of Turku, Finland}.\\
        Emails: \textsuperscript{1}\{xianjia.yu, jopequ, tovewe\}@utu.fi\\[+6pt]
    }
}

\begin{document}

\maketitle
\thispagestyle{empty}
\pagestyle{empty}



\begin{abstract}%
    \label{sec:abstract}%
    %
    The role of deep learning (DL) in robotics has significantly deepened over the last decade. Intelligent robotic systems today are highly connected systems that rely on DL for a variety of perception, control and other tasks. At the same time, autonomous robots are being increasingly deployed as part of fleets, with collaboration among robots becoming a more relevant factor. From the perspective of collaborative learning, federated learning (FL) enables continuous training of models in a distributed, privacy-preserving way.
    This paper
    focuses on vision-based obstacle avoidance for mobile robot navigation. On this basis, we explore the potential of FL for distributed systems of mobile robots enabling continuous learning via the engagement of robots in both simulated and real-world scenarios. 
    We extend previous works by studying the performance of different image classifiers for FL, compared to centralized, cloud-based learning
    %
    with a priori aggregated data. 
    We also introduce an approach to continuous learning from mobile robots with extended sensor suites able to provide automatically labelled data while they are completing other tasks. We show that higher accuracies can be achieved by training the models in both simulation and reality, enabling continuous updates to deployed models.
    %
\end{abstract}

\IEEEpeerreviewmaketitle


\section{Introduction}
\label{sec:introduction}

Vision-based obstacle avoidance is 
an important enabler for widespread autonomous robot navigation and could also help visually handicapped citizens~\cite{mendes2018assis}. In general, vision sensors have been having an increasing role in intelligent robotic systems, as advances in deep learning (DL) have led to new state-of-the-art solutions mobile navigation, human-like walking, teaching through demonstration, and collaborative automation, among others in areas such as advanced perception and intelligent control~\cite{pierson2017deep, zhao2020sim, queralta2020collaborative}.

In this paper, we extend the previous work~\cite{yu2022federated} on federated learning for vision-based obstacle avoidance in mobile robots towards analyzing different architectures and, most importantly, introducing an strategy for continuous learning from both simulated and real robots carrying out different missions requiring autonomous navigation. There are a number of works in the literature showing the potential of DL for vision-based obstacle avoidance~\cite{gaya2016vision}, and works integrating other techniques such multi-view structure-from-motion for enhanced performance~\cite{yang2017obstacle}. However, in addition to the use of DL in~\cite{yu2022federated}, the methods are also validated in the real world, a federated learning (FL) framework is introduced, and the effects of learning through heterogeneous environments are analyzed.

Federated learning enables sharing knowledge without transferring raw data. Therefore, FL allows for privacy-preserving distributed learning, which can then be enhanced by other technologies, such as distributed ledger technologies~\cite{xianjia2021federated}, has been utilized in multiple domains in robotics and autonomous system~\cite{kishor2022communication}.
To the best of our knowledge, this is the first work to introduce an strategy for continuous learning together with federated learning and validating it in the real world.
In addition to real-world robots, we also utilize simulations to obtain enough data for training the DL models. Due to the proliferation of photorealistic simulators in robotics fields, studies are increasingly relying on these simulators to supplement data collection in situations where we cannot reach or collect sufficient data. In addition, through these types of simulators, the deployment of robotic and autonomous systems can be more accessible. In this way, we believe it may be beneficial for real-world vision-based obstacle avoidance in the real world. Throughout their lives, humans and animals can acquire, fine-tune, and transfer knowledge and skills. It is instrumental and interesting for robots to have the same type of learning capabilities. To continually learn the model for obstacle avoidance like humans, we involved multiple robots performing other tasks in different places, including the simulator, into the FL-based lifelong learning system for data collection, model training, and model sharing. We compare the performance of both AlexNet and ResNet18 with centralized and federated learning approaches to model training. We evaluate the model on different synthetic and real-world datasets, and train the models on combinations of different subsets to better study the effect of environment heterogeneity during the training phase. New data is also acquired with a Clearpath Husky mobile robot while it is operating other tasks and autonomously navigating a new indoors environment. The new data is automatically labelled with additional sensor suited. This opens the door to wider usage of heterogeneous robot fleets where robots with additional sensor capabilities can generate labelled data to train models able to reproduce their autonomous behaviour with more limited sensors, mainly cameras.


Aided by AI, especially DL, each robot can have a model representing the environment based on their situated awareness. Different robots may have the limitation of detecting the environment due to their limited resources. It is of great necessity to have a collective model to share their knowledge about the environment. Instead of an individual robot, in a multi-robot system, multiple robots, situated in various whereabouts, collaboratively performing particular tasks is more efficient and of high success rate in heterogeneous environments including unknown ones~\cite{olcay2020collective}. However, because of the limits of scenarios where vision-based obstacle avoidance is applied, it is still challenging for DL to collect enough visual inputs for collaborative learning and share them with other agents for privacy and security reasons. To address these concerns, we proposed an approach that included FL, Sim-to-Real (Sim2Real) via a photorealistic simulator, and Lifelong Learning. 

In summary,  we investigated the possibility of federated and continuous learning within hybrid teams of simulated and real robotic agents in this work. We then evaluate the performance benefits of such an approach over offline learning or learning from more limited data sources. First, we evaluate two different deep obstacle avoidance neural networks with both synthetic and real-world data. For both of the architectures 
a FL-based knowledge sharing method (where the locally trained models are fused) is compared to a centralized training approach (where raw data needs to be aggregated before the training starts). 
Both models are validated with data from the photorealistic simulator and real-world environment separately. 
Second, we analyzed the sim-to-real performance of two different deep obstacle avoidance models generated by FL methods, with our results showing that FL outperforms the centralized data aggregation methods. Third, we integrate lidar-based navigation for automated labelled data gathering. We implemented an online FL-based visual obstacle avoidance system both in a simulator and real-world environment. With such a system, we can continuously collect data from obstacles and free paths and train the model while the robots operate other tasks.
The rest of this paper is organized as follows. In Section II we review related works in FL for autonomous robots, sim-to-real transfer and lifelong robot learning. The methodology followed for obtaining the results reported in the manuscript is then introduced in Section III, with Section IV delving into the actual experimental results. Finally, section V concludes the paper and outlines future work directions.




\section{Related Work} \label{sec:related_work}


Multi-robot collaboration and DL for robotics have both played an increasingly important role in multiple robotic applications~\cite{queralta2020collaborative}. However, most of the work to date in learning from real-world experiences, sim-to-real transfer, and continuous learning, has been dedicated to reinforcement learning~\cite{zhao2020sim, liu2019lifelong} and robotic manipulation~\cite{zhao2020ubiquitous}. Within the possibilities to achieve collaborative learning, one of the most straightforward approaches is cloud-based centralized learning~\cite{liu2021peer}, with a server where data is aggregated and training occurs at once or in batches, but in an offline manner. Federated learning, in contrast, offers a distributed solution to collaborative learning where, in turn, the process is privacy-preserving (raw data can be processed locally). It also allows for management of networking resources (choosing when and how to transfer the models) through distributed computation at the edge~\cite{imteaj2021survey}.


In the area of lifelong learning and continuous learning for robotics, the literature has been in general dedicated to simulation environments only. While AI advances the intelligence of robots in various subfields of robotics, numerous challenges remain with real-world application, particularly in an unknown environment. Among these limitations are a lack of open-ended learning about item categories and scenes, the absence of a global navigation map, and the occurrence of object collisions, particularly in dynamic environments. With the assistance of Lifelong learning, the robot should be able to learn new object categories continuously and comprehend affordances, especially for robots manipulating tasks from a small number of on-site training examples~\cite{kasaei2021state}. Lifelong learning for navigation is a good application of lifelong learning. Compared with the traditional planner in navigation, this learning approach can improve the navigation performance based on its own experience and retain the ability after learning new ones~\cite{liu2021lifelong}.




In terms of deep learning vision-based obstacle avoidance, %
%
this has been a researched topic partly owing to the 
higher degree of maturity of DL methods for vision data in comparison to other modalities such as lidar data~\cite{li2020deep, xianjia2022analyzing}. Cameras are also more widely available and are relatively inexpensive sensors, while the methods can be generalized to various environments owing to the lack of the need for specific geometric or topological features. For example, in~\cite{8917687}, autonomous aerial robots are deployed with minimal knowledge of the environment and obstacle avoidance is achieved through a single monocular camera. With models trained through deep reinforcement learning (DRL), noise to data in the training phase can bring increased resilience to changes in the environments~\cite{9636512}, a method that is typical in other DRL approaches~\cite{zhao2020sim, zhao2020towards}. In this paper, we also exploit heterogeneity of environments to show how a wider variety of the training data brings performance improvements to the DL models, especially when they are trained with a FL approach.




In the domain of sim-to-real to transfer for DL policies driving autonomous behaviour in robotic system, the most widely studied learning algorithms are DRL approaches~\cite{zhao2020sim}. Sim-to-real transfer with DRL has been particularly studied in depth in problems involving dexterous manipulation~\cite{zhao2020towards}. Even in this area, recent works in the literature also show the potential of DRL for vision-based obstacle avoidance~\cite{zhang2020sim2real, zhang2021sim2real}. The authors of these works demonstrate the effectiveness of the methods even when the robots encounter in the real world new environments and objects.

\section{Methodology}\label{sec:methodology}

This section outlines the robot platform utilized in this experiment, two different deep neural networks to do vision-based obstacle avoidance for further validation of FL performance, and navigation settings for continual learning. 

\subsection{Data collection for FL}

The details of the data gathering, including the usage of the photorealistic simulator Nvidia Isaac Sim, the environment settings, and data distributions, 
are detailed in~\cite{yu2022federated}, from where we reuse the base simulation and training datasets.
The datasets from~\cite{yu2022federated}
include data from three distinct scenarios in NVIDIA Isaac Sim and from Jetbot robots 
deployed in three real-world rooms to train and validate the vision-based obstacle avoidance models. In this work, the datasets from the simulator are represented as $\mathcal{S}_{i,\:i\in\{0,1,2\}}$where $i$ indicates the simulated environments, including a hospital, office, and warehouse.
The three real-world datasets are denoted as  $\mathcal{R}_{i,\:i\in\{0,1,2\}}$ where $i$ indicates office spaces, hallways, and laboratory environments, respectively. Additional datasets acquired specifically for this work are introduced in the relevant sections, with Husky training data referred to as $\mathcal{H}^{S}$ or $\mathcal{H}^{R}$ for the simulated and real robots, respectively.

Regarding data training hardware, in this work, we utilized a Lambda Vector workstation equipped with two RTX 3080 GPU cards and a 24-core AMD Threadripper 3960X processor to train our models for vision-based obstacle avoidance.

\subsection{Vision-based obstacle avoidance models}

We trained two distinct types of DL models to ensure FL's performance in visual obstacle avoidance and assess how performance differs. These two deep convolutional neural networks (CNN) are vision-based obstacle classifiers for two classes that define whether the environment ahead is \textit{blocked} or \textit{free} for the robot to navigate. Owing to the relatively low level of complexity of the classification task and the size of potential datasets for such tasks, we have selected the AlexNet~\cite{krizhevsky2012imagenet} and ResNet18~\cite{he2016deep} architectures as appropriate for such binary classification task. AlexNet and ResNet are both commonly exploited backbones for conducting various tasks across multiple domains.

These two models are generic deep learning models designed to aid robots in discerning between various types of barriers in heterogeneous situations. This strategy contrasts with other possibilities, such as object detection or semantic segmentation (e.g., segmenting free floor from objects and walls). The chosen approach enables us to concentrate on examining the performance of a federated learning approach and its capacity for sim-to-real transfer rather than on developing a specific obstacle avoidance strategy, which is the study's primary purpose.



\subsection{Proposed FL Based Lifelong Learning Obstacle Avoidance}

By incorporating lidar into the Husky navigation system, we developed a straightforward FL-based lifelong learning system for visual obstacle avoidance. We used lidar to differentiate between blocked and free space of visual data while performing specified navigation tasks. After collecting sufficient data, the robots will train the models independently and send them to the server for aggregation into a global model. The model can then be deployed to another type of robot for obstacle avoidance performance evaluation.

\begin{figure}[t]
    \centering
    \includegraphics[width=0.48\textwidth]{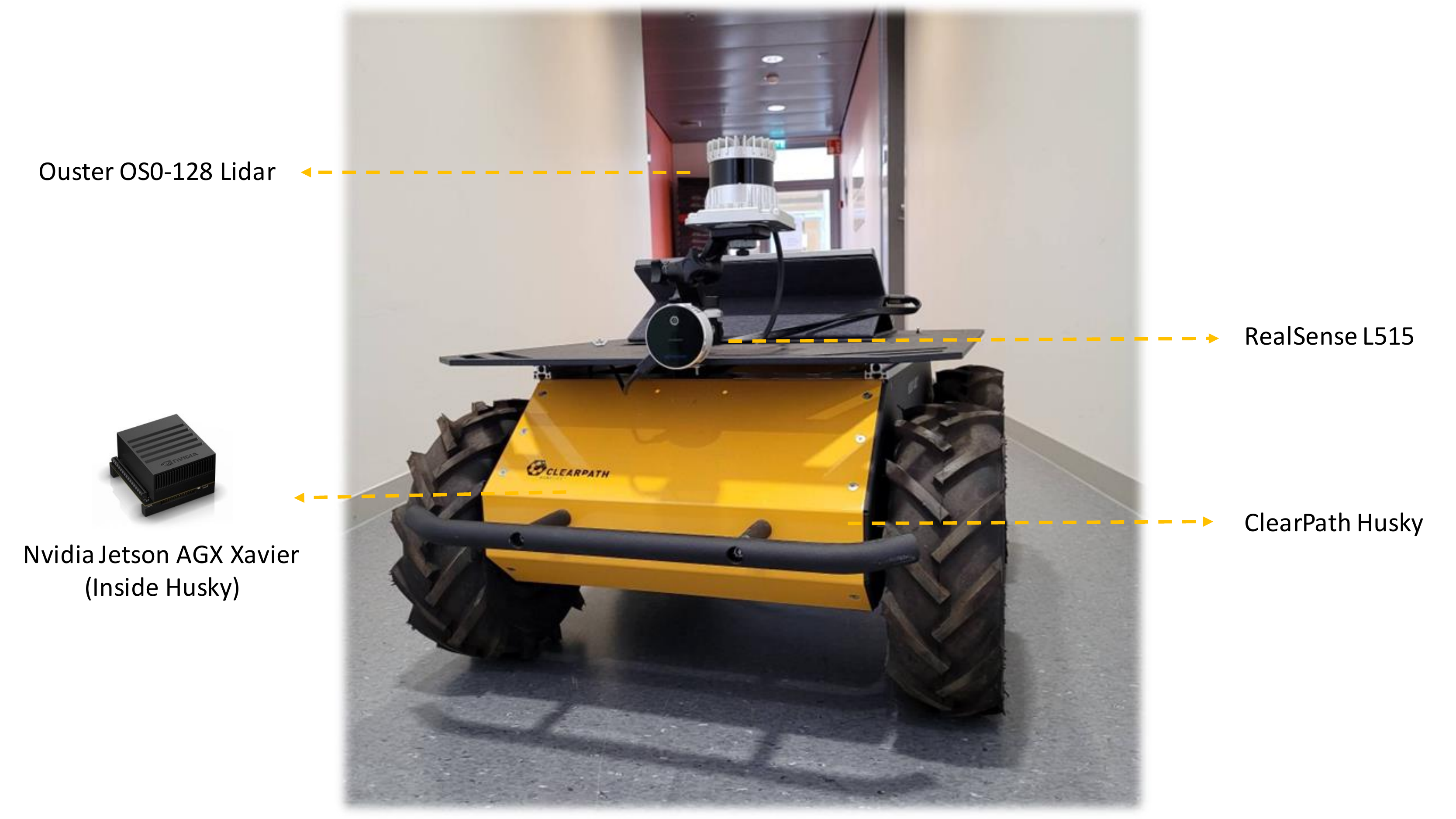}
    \caption{The customized Clearpath Husky platform.}
    \label{fig:husky}
\end{figure}

Regarding the experimental environments, We implemented this with a Clearpath Husky~\cref{fig:husky} robot both in a simulator and real-world environments. Our customized Husky robot platform is equipped with one Ouster OS0-128 lidar for measuring distances to objects when performing autonomous navigation tasks. When the distances are lower to a certain threshold, an Intel Realsense L515 camera will be triggered to categorize the images as obstacles included or free space continuously. Once a sufficient number of images are collected, we will train local models based on these data, fuse them to be a global model by FL methods, and then apply the global model to a real-world robot obstacle avoidance operation.

It's worth emphasizing that we used simulated data to help the continuous learning process, as obtaining real-world data is not always straightforward. In the following section, we evaluate the FL fused model's real-world deployment performance using either simulated or real-world data.

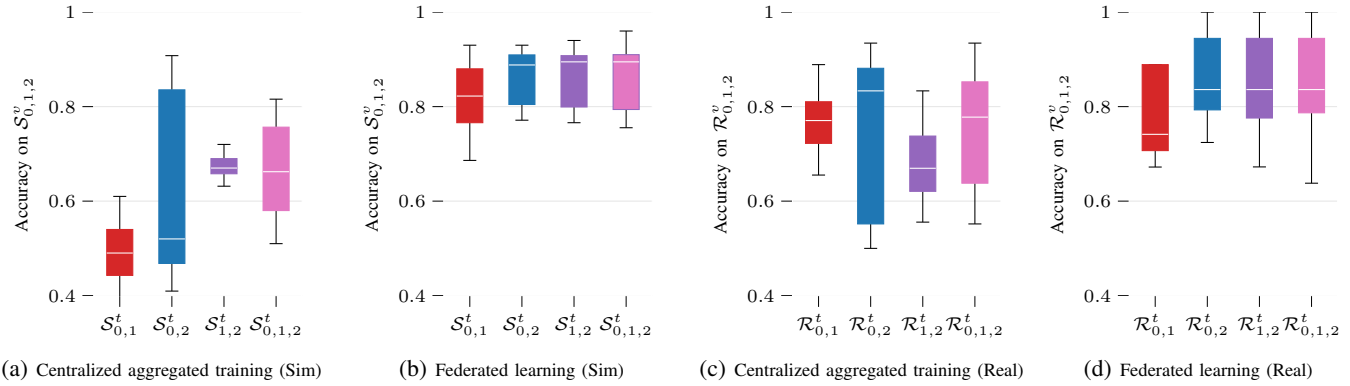
\begin{figure*}[t]
    \centering
    \begin{subfigure}{.24\textwidth}
        \centering
        \setlength{\figurewidth}{\textwidth}
        \setlength{\figureheight}{1.23\textwidth}
        \scriptsize{
\begin{tikzpicture}

\definecolor{color0}{rgb}{0.1,0.1,0.1}
\definecolor{color1}{rgb}{0.83921568627451,0.152941176470588,0.156862745098039}
\definecolor{color2}{rgb}{0.12156862745098,0.466666666666667,0.705882352941177}
\definecolor{color3}{rgb}{0.580392156862745,0.403921568627451,0.741176470588235}
\definecolor{color4}{rgb}{0.890196078431372,0.466666666666667,0.76078431372549}

\begin{axis}[
    height=\figureheight,
    width=\figurewidth,
    legend style={fill opacity=0.8, draw opacity=1, text opacity=1, draw=white!80!black},
    axis line style = {white},
    tick align=outside,
    tick pos=left,
    x grid style={white!69.0196078431373!black},
    xmin=0.5, xmax=4.5,
    xtick style={color=black},
    xtick={1,2,3,4},
    xticklabels={$\mathcal{S}_{0,1}^t$, $\mathcal{S}_{0,2}^t$, $\mathcal{S}_{1,2}^t$ , $\mathcal{S}_{0,1,2}^t$},
    y grid style={white!90!black},
    ylabel={Accuracy on $\mathcal{S}_{0,1,2}^v$},
    ymajorgrids,
    ymin=0.4, ymax=1.0,
    ytick style={color=black},
    scaled y ticks = false,
]
\addplot [black, forget plot]
table {%
1 0.442819155752659
1 0.382978737354279
};
\addplot [black, forget plot]
table {%
1 0.540000021457672
1 0.610000014305115
};
\addplot [black, forget plot]
table {%
0.875 0.382978737354279
1.125 0.382978737354279
};
\addplot [black, forget plot]
table {%
0.875 0.610000014305115
1.125 0.610000014305115
};
\addplot [black, forget plot]
table {%
2 0.468085110187531
2 0.40957447886467
};
\addplot [black, forget plot]
table {%
2 0.835526287555695
2 0.907894730567932
};
\addplot [black, forget plot]
table {%
1.875 0.40957447886467
2.125 0.40957447886467
};
\addplot [black, forget plot]
table {%
1.875 0.907894730567932
2.125 0.907894730567932
};
\addplot [black, forget plot]
table {%
3 0.657894730567932
3 0.631578922271729
};
\addplot [black, forget plot]
table {%
3 0.689999997615814
3 0.720000028610229
};
\addplot [black, forget plot]
table {%
2.875 0.631578922271729
3.125 0.631578922271729
};
\addplot [black, forget plot]
table {%
2.875 0.720000028610229
3.125 0.720000028610229
};
\addplot [black, forget plot]
table {%
4 0.579999983310699
4 0.509999990463257
};
\addplot [black, forget plot]
table {%
4 0.756578922271729
4 0.815789461135864
};
\addplot [black, forget plot]
table {%
3.875 0.509999990463257
4.125 0.509999990463257
};
\addplot [black, forget plot]
table {%
3.875 0.815789461135864
4.125 0.815789461135864
};
\path [draw=color1, fill=color1]
(axis cs:0.75,0.442819155752659)
--(axis cs:1.25,0.442819155752659)
--(axis cs:1.25,0.540000021457672)
--(axis cs:0.75,0.540000021457672)
--(axis cs:0.75,0.442819155752659)
--cycle;
\path [draw=color2, fill=color2]
(axis cs:1.75,0.468085110187531)
--(axis cs:2.25,0.468085110187531)
--(axis cs:2.25,0.835526287555695)
--(axis cs:1.75,0.835526287555695)
--(axis cs:1.75,0.468085110187531)
--cycle;
\path [draw=color3, fill=color3]
(axis cs:2.75,0.657894730567932)
--(axis cs:3.25,0.657894730567932)
--(axis cs:3.25,0.689999997615814)
--(axis cs:2.75,0.689999997615814)
--(axis cs:2.75,0.657894730567932)
--cycle;
\path [draw=color4, fill=color4]
(axis cs:3.75,0.579999983310699)
--(axis cs:4.25,0.579999983310699)
--(axis cs:4.25,0.756578922271729)
--(axis cs:3.75,0.756578922271729)
--(axis cs:3.75,0.579999983310699)
--cycle;
\addplot [white, forget plot]
table {%
0.75 0.490131571888924
1.25 0.490131571888924
};
\addplot [white, forget plot]
table {%
1.75 0.519999980926514
2.25 0.519999980926514
};
\addplot [white, forget plot]
table {%
2.75 0.670000016689301
3.25 0.670000016689301
};
\addplot [white, forget plot]
table {%
3.75 0.662234038114548
4.25 0.662234038114548
};
\end{axis}

\end{tikzpicture}}
        \caption{\scriptsize{Centralized aggregated training (Sim)}}
        \label{resnet_sim_acc_mixed}
    \end{subfigure}
    \hfill
    \begin{subfigure}{.24\textwidth}
      \centering
        \setlength{\figurewidth}{\textwidth}
        \setlength{\figureheight}{1.23\textwidth}
        \scriptsize{
\begin{tikzpicture}

\definecolor{color0}{rgb}{0.1,0.1,0.1}
\definecolor{color1}{rgb}{0.83921568627451,0.152941176470588,0.156862745098039}
\definecolor{color2}{rgb}{0.12156862745098,0.466666666666667,0.705882352941177}
\definecolor{color3}{rgb}{0.580392156862745,0.403921568627451,0.741176470588235}
\definecolor{color4}{rgb}{0.890196078431372,0.466666666666667,0.76078431372549}
\begin{axis}[
     height=\figureheight,
     width=\figurewidth,
    legend style={fill opacity=0.8, draw opacity=1, text opacity=1, draw=white!80!black},
    axis line style = {white},
    tick align=outside,
    tick pos=left,
    x grid style={white!69.0196078431373!black},
    xmin=0.5, xmax=4.5,
    xtick style={color=black},
    xtick={1,2,3,4},
    xticklabels={$\mathcal{S}_{0,1}^t$, $\mathcal{S}_{0,2}^t$, $\mathcal{S}_{1,2}^t$ , $\mathcal{S}_{0,1,2}^t$},
    y grid style={white!90!black},
    ylabel={Accuracy on $\mathcal{S}_{0,1,2}^v$},
    ymajorgrids,
    ymin=0.4, ymax=1.0,
    ytick style={color=black},
]
\addplot [black, forget plot]
table {%
1 0.765957474708557
1 0.686170220375061
};
\addplot [black, forget plot]
table {%
1 0.879999995231628
1 0.930000007152557
};
\addplot [black, forget plot]
table {%
0.875 0.686170220375061
1.125 0.686170220375061
};
\addplot [black, forget plot]
table {%
0.875 0.930000007152557
1.125 0.930000007152557
};
\addplot [black, forget plot]
table {%
2 0.804521277546883
2 0.771276593208313
};
\addplot [black, forget plot]
table {%
2 0.909473702311516
2 0.930000007152557
};
\addplot [black, forget plot]
table {%
1.875 0.771276593208313
2.125 0.771276593208313
};
\addplot [black, forget plot]
table {%
1.875 0.930000007152557
2.125 0.930000007152557
};
\addplot [black, forget plot]
table {%
3 0.799202144145966
3 0.765957474708557
};
\addplot [black, forget plot]
table {%
3 0.907894730567932
3 0.939999997615814
};
\addplot [black, forget plot]
table {%
2.875 0.765957474708557
3.125 0.765957474708557
};
\addplot [black, forget plot]
table {%
2.875 0.939999997615814
3.125 0.939999997615814
};
\addplot [black, forget plot]
table {%
4 0.793882980942726
4 0.755319178104401
};
\addplot [black, forget plot]
table {%
4 0.910000026226044
4 0.959999978542328
};
\addplot [black, forget plot]
table {%
3.875 0.755319178104401
4.125 0.755319178104401
};
\addplot [black, forget plot]
table {%
3.875 0.959999978542328
4.125 0.959999978542328
};
\path [draw=color1, fill=color1]
(axis cs:0.75,0.765957474708557)
--(axis cs:1.25,0.765957474708557)
--(axis cs:1.25,0.879999995231628)
--(axis cs:0.75,0.879999995231628)
--(axis cs:0.75,0.765957474708557)
--cycle;
\path [draw=color2, fill=color2]
(axis cs:1.75,0.804521277546883)
--(axis cs:2.25,0.804521277546883)
--(axis cs:2.25,0.909473702311516)
--(axis cs:1.75,0.909473702311516)
--(axis cs:1.75,0.804521277546883)
--cycle;
\path [draw=color3, fill=color3]
(axis cs:2.75,0.799202144145966)
--(axis cs:3.25,0.799202144145966)
--(axis cs:3.25,0.907894730567932)
--(axis cs:2.75,0.907894730567932)
--(axis cs:2.75,0.799202144145966)
--cycle;
\path [draw=color3, fill=color4]
(axis cs:3.75,0.793882980942726)
--(axis cs:4.25,0.793882980942726)
--(axis cs:4.25,0.910000026226044)
--(axis cs:3.75,0.910000026226044)
--(axis cs:3.75,0.793882980942726)
--cycle;
\addplot [white, forget plot]
table {%
0.75 0.822368443012238
1.25 0.822368443012238
};
\addplot [white, forget plot]
table {%
1.75 0.888157904148102
2.25 0.888157904148102
};
\addplot [white, forget plot]
table {%
2.75 0.89473682641983
3.25 0.89473682641983
};
\addplot [white, forget plot]
table {%
3.75 0.89473682641983
4.25 0.89473682641983
};
\end{axis}

\end{tikzpicture}}
        \caption{\scriptsize{Federated learning (Sim)}}
      \label{fig:resnet_sim_acc_fused}
    \end{subfigure}
    \hfill
    \begin{subfigure}{.24\textwidth}
      \centering
        \setlength{\figurewidth}{\textwidth}
        \setlength{\figureheight}{1.23\textwidth}
        \scriptsize{
\begin{tikzpicture}

\definecolor{color0}{rgb}{0.1,0.1,0.1}
\definecolor{color1}{rgb}{0.83921568627451,0.152941176470588,0.156862745098039}
\definecolor{color2}{rgb}{0.12156862745098,0.466666666666667,0.705882352941177}
\definecolor{color3}{rgb}{0.580392156862745,0.403921568627451,0.741176470588235}
\definecolor{color4}{rgb}{0.890196078431372,0.466666666666667,0.76078431372549}

\begin{axis}[
     height=\figureheight,
     width=\figurewidth,
    legend style={fill opacity=0.8, draw opacity=1, text opacity=1, draw=white!80!black},
    axis line style = {white},
    tick align=outside,
    tick pos=left,
    x grid style={white!69.0196078431373!black},
    xmin=0.5, xmax=4.5,
    xtick style={color=black},
    xtick={1,2,3,4},
    xticklabels={$\mathcal{R}_{0,1}^t$, $\mathcal{R}_{0,2}^t$, $\mathcal{R}_{1,2}^t$ , $\mathcal{R}_{0,1,2}^t$},
    y grid style={white!90!black},
    ylabel={Accuracy on $\mathcal{R}_{0,1,2}^v$},
    ymajorgrids,
    ymin=0.4, ymax=1.0,
    ytick style={color=black},
]
\addplot [black, forget plot]
table {%
1 0.722222208976746
1 0.655172407627106
};
\addplot [black, forget plot]
table {%
1 0.810344815254211
1 0.888888895511627
};
\addplot [black, forget plot]
table {%
0.875 0.655172407627106
1.125 0.655172407627106
};
\addplot [black, forget plot]
table {%
0.875 0.888888895511627
1.125 0.888888895511627
};
\addplot [black, forget plot]
table {%
2 0.551724135875702
2 0.5
};
\addplot [black, forget plot]
table {%
2 0.881147548556328
2 0.934426248073578
};
\addplot [black, forget plot]
table {%
1.875 0.5
2.125 0.5
};
\addplot [black, forget plot]
table {%
1.875 0.934426248073578
2.125 0.934426248073578
};
\addplot [black, forget plot]
table {%
3 0.620689630508423
3 0.555555582046509
};
\addplot [black, forget plot]
table {%
3 0.737704932689667
3 0.833333313465118
};
\addplot [black, forget plot]
table {%
2.875 0.555555582046509
3.125 0.555555582046509
};
\addplot [black, forget plot]
table {%
2.875 0.833333313465118
3.125 0.833333313465118
};
\addplot [black, forget plot]
table {%
4 0.637931048870087
4 0.551724135875702
};
\addplot [black, forget plot]
table {%
4 0.852459013462067
4 0.934426248073578
};
\addplot [black, forget plot]
table {%
3.875 0.551724135875702
4.125 0.551724135875702
};
\addplot [black, forget plot]
table {%
3.875 0.934426248073578
4.125 0.934426248073578
};
\path [draw=color1, fill=color1]
(axis cs:0.75,0.722222208976746)
--(axis cs:1.25,0.722222208976746)
--(axis cs:1.25,0.810344815254211)
--(axis cs:0.75,0.810344815254211)
--(axis cs:0.75,0.722222208976746)
--cycle;
\path [draw=color2, fill=color2]
(axis cs:1.75,0.551724135875702)
--(axis cs:2.25,0.551724135875702)
--(axis cs:2.25,0.881147548556328)
--(axis cs:1.75,0.881147548556328)
--(axis cs:1.75,0.551724135875702)
--cycle;
\path [draw=color3, fill=color3]
(axis cs:2.75,0.620689630508423)
--(axis cs:3.25,0.620689630508423)
--(axis cs:3.25,0.737704932689667)
--(axis cs:2.75,0.737704932689667)
--(axis cs:2.75,0.620689630508423)
--cycle;
\path [draw=color4, fill=color4]
(axis cs:3.75,0.637931048870087)
--(axis cs:4.25,0.637931048870087)
--(axis cs:4.25,0.852459013462067)
--(axis cs:3.75,0.852459013462067)
--(axis cs:3.75,0.637931048870087)
--cycle;
\addplot [white, forget plot]
table {%
0.75 0.770491778850555
1.25 0.770491778850555
};
\addplot [white, forget plot]
table {%
1.75 0.833333313465118
2.25 0.833333313465118
};
\addplot [white, forget plot]
table {%
2.75 0.669540226459503
3.25 0.669540226459503
};
\addplot [white, forget plot]
table {%
3.75 0.777777791023254
4.25 0.777777791023254
};
\end{axis}

\end{tikzpicture}}
        \caption{\scriptsize{Centralized aggregated training (Real)}}
      \label{fig:resnet_real_acc_mixed}
    \end{subfigure}
        \begin{subfigure}{.24\textwidth}
      \centering
        \setlength{\figurewidth}{\textwidth}
        \setlength{\figureheight}{1.23\textwidth}
        \scriptsize{
\begin{tikzpicture}

\definecolor{color0}{rgb}{0.1,0.1,0.1}
\definecolor{color1}{rgb}{0.83921568627451,0.152941176470588,0.156862745098039}
\definecolor{color2}{rgb}{0.12156862745098,0.466666666666667,0.705882352941177}
\definecolor{color3}{rgb}{0.580392156862745,0.403921568627451,0.741176470588235}
\definecolor{color4}{rgb}{0.890196078431372,0.466666666666667,0.76078431372549}

\begin{axis}[
     height=\figureheight,
     width=\figurewidth,
    legend style={fill opacity=0.8, draw opacity=1, text opacity=1, draw=white!80!black},
    axis line style = {white},
    tick align=outside,
    tick pos=left,
    x grid style={white!69.0196078431373!black},
    xmin=0.5, xmax=4.5,
    xtick style={color=black},
    xtick={1,2,3,4},
    xticklabels={$\mathcal{R}_{0,1}^t$, $\mathcal{R}_{0,2}^t$, $\mathcal{R}_{1,2}^t$ , $\mathcal{R}_{0,1,2}^t$},
    y grid style={white!90!black},
    ylabel={Accuracy on $\mathcal{R}_{0,1,2}^v$},
    ymajorgrids,
    ymin=0.4, ymax=1.0,
    ytick style={color=black},
    ]
\addplot [black, forget plot]
table {%
1 0.706896543502808
1 0.6721311211586
};
\addplot [black, forget plot]
table {%
1 0.888888895511627
1 0.888888895511627
};
\addplot [black, forget plot]
table {%
0.875 0.6721311211586
1.125 0.6721311211586
};
\addplot [black, forget plot]
table {%
0.875 0.888888895511627
1.125 0.888888895511627
};
\addplot [black, forget plot]
table {%
2 0.793103456497192
2 0.724137902259827
};
\addplot [black, forget plot]
table {%
2 0.944444417953491
2 1
};
\addplot [black, forget plot]
table {%
1.875 0.724137902259827
2.125 0.724137902259827
};
\addplot [black, forget plot]
table {%
1.875 1
2.125 1
};
\addplot [black, forget plot]
table {%
3 0.775862097740173
3 0.672413766384125
};
\addplot [black, forget plot]
table {%
3 0.944444417953491
3 1
};
\addplot [black, forget plot]
table {%
2.875 0.672413766384125
3.125 0.672413766384125
};
\addplot [black, forget plot]
table {%
2.875 1
3.125 1
};
\addplot [black, forget plot]
table {%
4 0.786885261535645
4 0.637931048870087
};
\addplot [black, forget plot]
table {%
4 0.944444417953491
4 1
};
\addplot [black, forget plot]
table {%
3.875 0.637931048870087
4.125 0.637931048870087
};
\addplot [black, forget plot]
table {%
3.875 1
4.125 1
};
\path [draw=color1, fill=color1]
(axis cs:0.75,0.706896543502808)
--(axis cs:1.25,0.706896543502808)
--(axis cs:1.25,0.888888895511627)
--(axis cs:0.75,0.888888895511627)
--(axis cs:0.75,0.706896543502808)
--cycle;
\path [draw=color2, fill=color2]
(axis cs:1.75,0.793103456497192)
--(axis cs:2.25,0.793103456497192)
--(axis cs:2.25,0.944444417953491)
--(axis cs:1.75,0.944444417953491)
--(axis cs:1.75,0.793103456497192)
--cycle;
\path [draw=color3, fill=color3]
(axis cs:2.75,0.775862097740173)
--(axis cs:3.25,0.775862097740173)
--(axis cs:3.25,0.944444417953491)
--(axis cs:2.75,0.944444417953491)
--(axis cs:2.75,0.775862097740173)
--cycle;
\path [draw=color4, fill=color4]
(axis cs:3.75,0.786885261535645)
--(axis cs:4.25,0.786885261535645)
--(axis cs:4.25,0.944444417953491)
--(axis cs:3.75,0.944444417953491)
--(axis cs:3.75,0.786885261535645)
--cycle;
\addplot [white, forget plot]
table {%
0.75 0.74137932062149
1.25 0.74137932062149
};
\addplot [white, forget plot]
table {%
1.75 0.836065590381622
2.25 0.836065590381622
};
\addplot [white, forget plot]
table {%
2.75 0.836065590381622
3.25 0.836065590381622
};
\addplot [white, forget plot]
table {%
3.75 0.836065590381622
4.25 0.836065590381622
};
\end{axis}

\end{tikzpicture}}
        \caption{\scriptsize{Federated learning (Real)}}
      \label{fig:resnet_real_acc_fused}
    \end{subfigure}
    \caption{Accuracy of models obtained through centralized learning with aggregated data or federated learning with fused local models based on ResNet18. These results are trained ($^t$) and validated ($^v$) with respective simulation datasets ($\mathcal{S}_i^t$, $\mathcal{S}_i^v$) and real datasets ($\mathcal{R}_i^t$, $\mathcal{R}_i^v$) independently.}
    \label{fig:resnet_accuracy_results}
\end{figure*}

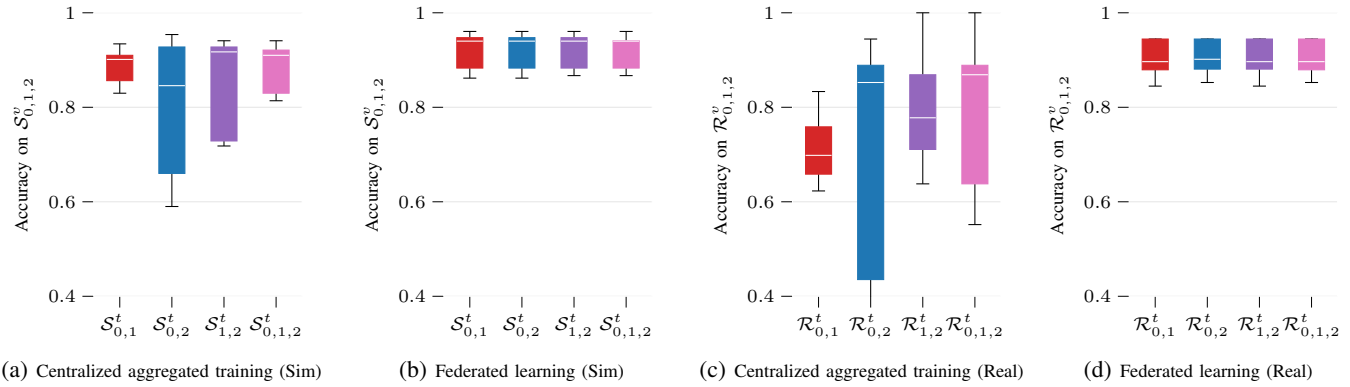
\begin{figure*}[t]
    \centering
    \begin{subfigure}{.24\textwidth}
        \centering
        \setlength{\figurewidth}{\textwidth}
        \setlength{\figureheight}{1.23\textwidth}
        \scriptsize{
\begin{tikzpicture}

\definecolor{color0}{rgb}{1,0.498039215686275,0.0549019607843137}
\definecolor{color1}{rgb}{0.83921568627451,0.152941176470588,0.156862745098039}
\definecolor{color2}{rgb}{0.12156862745098,0.466666666666667,0.705882352941177}
\definecolor{color3}{rgb}{0.580392156862745,0.403921568627451,0.741176470588235}
\definecolor{color4}{rgb}{0.890196078431372,0.466666666666667,0.76078431372549}

\begin{axis}[
     height=\figureheight,
     width=\figurewidth,
    legend style={fill opacity=0.8, draw opacity=1, text opacity=1, draw=white!80!black},
    axis line style = {white},
    tick align=outside,
    tick pos=left,
    x grid style={white!69.0196078431373!black},
    xmin=0.5, xmax=4.5,
    xtick style={color=black},
    xtick={1,2,3,4},
    xticklabels={$\mathcal{S}_{0,1}^t$, $\mathcal{S}_{0,2}^t$, $\mathcal{S}_{1,2}^t$ , $\mathcal{S}_{0,1,2}^t$},
    y grid style={white!90!black},
    ylabel={Accuracy on $\mathcal{S}_{0,1,2}^v$},
    ymajorgrids,
    ymin=0.4, ymax=1.0,
    ytick style={color=black},
    scaled y ticks = false,
]
\addplot [black, forget plot]
table {%
1 0.856382966041565
1 0.829787254333496
};
\addplot [black, forget plot]
table {%
1 0.910000026226044
1 0.934210538864136
};
\addplot [black, forget plot]
table {%
0.875 0.829787254333496
1.125 0.829787254333496
};
\addplot [black, forget plot]
table {%
0.875 0.934210538864136
1.125 0.934210538864136
};
\addplot [black, forget plot]
table {%
2 0.660000026226044
2 0.589999973773956
};
\addplot [black, forget plot]
table {%
2 0.927631556987762
2 0.953947365283966
};
\addplot [black, forget plot]
table {%
1.875 0.589999973773956
2.125 0.589999973773956
};
\addplot [black, forget plot]
table {%
1.875 0.953947365283966
2.125 0.953947365283966
};
\addplot [black, forget plot]
table {%
3 0.728723406791687
3 0.718085110187531
};
\addplot [black, forget plot]
table {%
3 0.927631556987762
3 0.940789461135864
};
\addplot [black, forget plot]
table {%
2.875 0.718085110187531
3.125 0.718085110187531
};
\addplot [black, forget plot]
table {%
2.875 0.940789461135864
3.125 0.940789461135864
};
\addplot [black, forget plot]
table {%
4 0.829787254333496
4 0.813829779624939
};
\addplot [black, forget plot]
table {%
4 0.921052634716034
4 0.940789461135864
};
\addplot [black, forget plot]
table {%
3.875 0.813829779624939
4.125 0.813829779624939
};
\addplot [black, forget plot]
table {%
3.875 0.940789461135864
4.125 0.940789461135864
};
\path [draw=color1, fill=color1]
(axis cs:0.75,0.856382966041565)
--(axis cs:1.25,0.856382966041565)
--(axis cs:1.25,0.910000026226044)
--(axis cs:0.75,0.910000026226044)
--(axis cs:0.75,0.856382966041565)
--cycle;
\path [draw=color2, fill=color2]
(axis cs:1.75,0.660000026226044)
--(axis cs:2.25,0.660000026226044)
--(axis cs:2.25,0.927631556987762)
--(axis cs:1.75,0.927631556987762)
--(axis cs:1.75,0.660000026226044)
--cycle;
\path [draw=color3, fill=color3]
(axis cs:2.75,0.728723406791687)
--(axis cs:3.25,0.728723406791687)
--(axis cs:3.25,0.927631556987762)
--(axis cs:2.75,0.927631556987762)
--(axis cs:2.75,0.728723406791687)
--cycle;
\path [draw=color4, fill=color4]
(axis cs:3.75,0.829787254333496)
--(axis cs:4.25,0.829787254333496)
--(axis cs:4.25,0.921052634716034)
--(axis cs:3.75,0.921052634716034)
--(axis cs:3.75,0.829787254333496)
--cycle;
\addplot [white, forget plot]
table {%
0.75 0.901315808296204
1.25 0.901315808296204
};
\addplot [white, forget plot]
table {%
1.75 0.845744669437408
2.25 0.845744669437408
};
\addplot [white, forget plot]
table {%
2.75 0.917236864566803
3.25 0.917236864566803
};
\addplot [white, forget plot]
table {%
3.75 0.910000026226044
4.25 0.910000026226044
};
\end{axis}

\end{tikzpicture}}
        \caption{\scriptsize{Centralized aggregated training (Sim)}}
        \label{fig:alexnet_sim_acc_mixed}
    \end{subfigure}
    \hfill
    \begin{subfigure}{.24\textwidth}
      \centering
        \setlength{\figurewidth}{\textwidth}
        \setlength{\figureheight}{1.23\textwidth}
        \scriptsize{
\begin{tikzpicture}

\definecolor{color0}{rgb}{1,0.498039215686275,0.0549019607843137}
\definecolor{color1}{rgb}{0.83921568627451,0.152941176470588,0.156862745098039}
\definecolor{color2}{rgb}{0.12156862745098,0.466666666666667,0.705882352941177}
\definecolor{color3}{rgb}{0.580392156862745,0.403921568627451,0.741176470588235}
\definecolor{color4}{rgb}{0.890196078431372,0.466666666666667,0.76078431372549}

\begin{axis}[
     height=\figureheight,
     width=\figurewidth,
    legend style={fill opacity=0.8, draw opacity=1, text opacity=1, draw=white!80!black},
    axis line style = {white},
    tick align=outside,
    tick pos=left,
    x grid style={white!69.0196078431373!black},
    xmin=0.5, xmax=4.5,
    xtick style={color=black},
    xtick={1,2,3,4},
    xticklabels={$\mathcal{S}_{0,1}^t$, $\mathcal{S}_{0,2}^t$, $\mathcal{S}_{1,2}^t$ , $\mathcal{S}_{0,1,2}^t$},
    y grid style={white!90!black},
    ylabel={Accuracy on $\mathcal{S}_{0,1,2}^v$},
    ymajorgrids,
    ymin=0.4, ymax=1.0,
    ytick style={color=black},
]
\addplot [black, forget plot]
table {%
1 0.882978737354279
1 0.861702144145966
};
\addplot [black, forget plot]
table {%
1 0.947368443012238
1 0.960526287555695
};
\addplot [black, forget plot]
table {%
0.875 0.861702144145966
1.125 0.861702144145966
};
\addplot [black, forget plot]
table {%
0.875 0.960526287555695
1.125 0.960526287555695
};
\addplot [black, forget plot]
table {%
2 0.882978737354279
2 0.861702144145966
};
\addplot [black, forget plot]
table {%
2 0.947368443012238
2 0.960526287555695
};
\addplot [black, forget plot]
table {%
1.875 0.861702144145966
2.125 0.861702144145966
};
\addplot [black, forget plot]
table {%
1.875 0.960526287555695
2.125 0.960526287555695
};
\addplot [black, forget plot]
table {%
3 0.882978737354279
3 0.867021262645721
};
\addplot [black, forget plot]
table {%
3 0.947368443012238
3 0.960526287555695
};
\addplot [black, forget plot]
table {%
2.875 0.867021262645721
3.125 0.867021262645721
};
\addplot [black, forget plot]
table {%
2.875 0.960526287555695
3.125 0.960526287555695
};
\addplot [black, forget plot]
table {%
4 0.882978737354279
4 0.867021262645721
};
\addplot [black, forget plot]
table {%
4 0.940789461135864
4 0.960526287555695
};
\addplot [black, forget plot]
table {%
3.875 0.867021262645721
4.125 0.867021262645721
};
\addplot [black, forget plot]
table {%
3.875 0.960526287555695
4.125 0.960526287555695
};
\path [draw=color1, fill=color1]
(axis cs:0.75,0.882978737354279)
--(axis cs:1.25,0.882978737354279)
--(axis cs:1.25,0.947368443012238)
--(axis cs:0.75,0.947368443012238)
--(axis cs:0.75,0.882978737354279)
--cycle;
\path [draw=color2, fill=color2]
(axis cs:1.75,0.882978737354279)
--(axis cs:2.25,0.882978737354279)
--(axis cs:2.25,0.947368443012238)
--(axis cs:1.75,0.947368443012238)
--(axis cs:1.75,0.882978737354279)
--cycle;
\path [draw=color3, fill=color3]
(axis cs:2.75,0.882978737354279)
--(axis cs:3.25,0.882978737354279)
--(axis cs:3.25,0.947368443012238)
--(axis cs:2.75,0.947368443012238)
--(axis cs:2.75,0.882978737354279)
--cycle;
\path [draw=color4, fill=color4]
(axis cs:3.75,0.882978737354279)
--(axis cs:4.25,0.882978737354279)
--(axis cs:4.25,0.940789461135864)
--(axis cs:3.75,0.940789461135864)
--(axis cs:3.75,0.882978737354279)
--cycle;
\addplot [white, forget plot]
table {%
0.75 0.939999997615814
1.25 0.939999997615814
};
\addplot [white, forget plot]
table {%
1.75 0.939999997615814
2.25 0.939999997615814
};
\addplot [white, forget plot]
table {%
2.75 0.939999997615814
3.25 0.939999997615814
};
\addplot [white, forget plot]
table {%
3.75 0.939999997615814
4.25 0.939999997615814
};
\end{axis}

\end{tikzpicture}}
        \caption{\scriptsize{Federated learning (Sim)}}
      \label{fig:alexnet_sim_acc_fused}
    \end{subfigure}
    \hfill
    \begin{subfigure}{.24\textwidth}
      \centering
        \setlength{\figurewidth}{\textwidth}
        \setlength{\figureheight}{1.23\textwidth}
        \scriptsize{
\begin{tikzpicture}

\definecolor{color0}{rgb}{1,0.498039215686275,0.0549019607843137}
\definecolor{color1}{rgb}{0.83921568627451,0.152941176470588,0.156862745098039}
\definecolor{color2}{rgb}{0.12156862745098,0.466666666666667,0.705882352941177}
\definecolor{color3}{rgb}{0.580392156862745,0.403921568627451,0.741176470588235}
\definecolor{color4}{rgb}{0.890196078431372,0.466666666666667,0.76078431372549}

\begin{axis}[
     height=\figureheight,
     width=\figurewidth,
    legend style={fill opacity=0.8, draw opacity=1, text opacity=1, draw=white!80!black},
    axis line style = {white},
    tick align=outside,
    tick pos=left,
    x grid style={white!69.0196078431373!black},
    xmin=0.5, xmax=4.5,
    xtick style={color=black},
    xtick={1,2,3,4},
    xticklabels={$\mathcal{R}_{0,1}^t$, $\mathcal{R}_{0,2}^t$, $\mathcal{R}_{1,2}^t$ , $\mathcal{R}_{0,1,2}^t$},
    y grid style={white!90!black},
    ylabel={Accuracy on $\mathcal{R}_{0,1,2}^v$},
    ymajorgrids,
    ymin=0.4, ymax=1.0,
    ytick style={color=black},
]
\addplot [black, forget plot]
table {%
1 0.658469945192337
1 0.622950792312622
};
\addplot [black, forget plot]
table {%
1 0.75862067937851
1 0.833333313465118
};
\addplot [black, forget plot]
table {%
0.875 0.622950792312622
1.125 0.622950792312622
};
\addplot [black, forget plot]
table {%
0.875 0.833333313465118
1.125 0.833333313465118
};
\addplot [black, forget plot]
table {%
2 0.435344822704792
2 0.396551728248596
};
\addplot [black, forget plot]
table {%
2 0.888888895511627
2 0.944444417953491
};
\addplot [black, forget plot]
table {%
1.875 0.396551728248596
2.125 0.396551728248596
};
\addplot [black, forget plot]
table {%
1.875 0.944444417953491
2.125 0.944444417953491
};
\addplot [black, forget plot]
table {%
3 0.710727959871292
3 0.637931048870087
};
\addplot [black, forget plot]
table {%
3 0.868852436542511
3 1
};
\addplot [black, forget plot]
table {%
2.875 0.637931048870087
3.125 0.637931048870087
};
\addplot [black, forget plot]
table {%
2.875 1
3.125 1
};
\addplot [black, forget plot]
table {%
4 0.637931048870087
4 0.551724135875702
};
\addplot [black, forget plot]
table {%
4 0.888888895511627
4 1
};
\addplot [black, forget plot]
table {%
3.875 0.551724135875702
4.125 0.551724135875702
};
\addplot [black, forget plot]
table {%
3.875 1
4.125 1
};
\path [draw=color1, fill=color1]
(axis cs:0.75,0.658469945192337)
--(axis cs:1.25,0.658469945192337)
--(axis cs:1.25,0.75862067937851)
--(axis cs:0.75,0.75862067937851)
--(axis cs:0.75,0.658469945192337)
--cycle;
\path [draw=color2, fill=color2]
(axis cs:1.75,0.435344822704792)
--(axis cs:2.25,0.435344822704792)
--(axis cs:2.25,0.888888895511627)
--(axis cs:1.75,0.888888895511627)
--(axis cs:1.75,0.435344822704792)
--cycle;
\path [draw=color3, fill=color3]
(axis cs:2.75,0.710727959871292)
--(axis cs:3.25,0.710727959871292)
--(axis cs:3.25,0.868852436542511)
--(axis cs:2.75,0.868852436542511)
--(axis cs:2.75,0.710727959871292)
--cycle;
\path [draw=color4, fill=color4]
(axis cs:3.75,0.637931048870087)
--(axis cs:4.25,0.637931048870087)
--(axis cs:4.25,0.888888895511627)
--(axis cs:3.75,0.888888895511627)
--(axis cs:3.75,0.637931048870087)
--cycle;
\addplot [white, forget plot]
table {%
0.75 0.698275864124298
1.25 0.698275864124298
};
\addplot [white, forget plot]
table {%
1.75 0.852459013462067
2.25 0.852459013462067
};
\addplot [white, forget plot]
table {%
2.75 0.777777791023254
3.25 0.777777791023254
};
\addplot [white, forget plot]
table {%
3.75 0.868852436542511
4.25 0.868852436542511
};
\end{axis}

\end{tikzpicture}}
        \caption{\scriptsize{Centralized aggregated training (Real)}}
      \label{fig:alexnet_real_acc_mixed}
    \end{subfigure}
        \begin{subfigure}{.24\textwidth}
      \centering
        \setlength{\figurewidth}{\textwidth}
        \setlength{\figureheight}{1.23\textwidth}
        \scriptsize{
\begin{tikzpicture}

\definecolor{color0}{rgb}{1,0.498039215686275,0.0549019607843137}
\definecolor{color1}{rgb}{0.83921568627451,0.152941176470588,0.156862745098039}
\definecolor{color2}{rgb}{0.12156862745098,0.466666666666667,0.705882352941177}
\definecolor{color3}{rgb}{0.580392156862745,0.403921568627451,0.741176470588235}
\definecolor{color4}{rgb}{0.890196078431372,0.466666666666667,0.76078431372549}

\begin{axis}[
     height=\figureheight,
     width=\figurewidth,
    legend style={fill opacity=0.8, draw opacity=1, text opacity=1, draw=white!80!black},
    axis line style = {white},
    tick align=outside,
    tick pos=left,
    x grid style={white!69.0196078431373!black},
    xmin=0.5, xmax=4.5,
    xtick style={color=black},
    xtick={1,2,3,4},
    xticklabels={$\mathcal{R}_{0,1}^t$, $\mathcal{R}_{0,2}^t$, $\mathcal{R}_{1,2}^t$ , $\mathcal{R}_{0,1,2}^t$},
    y grid style={white!90!black},
    ylabel={Accuracy on $\mathcal{R}_{0,1,2}^v$},
    ymajorgrids,
    ymin=0.4, ymax=1.0,
    ytick style={color=black},
    ]
\addplot [black, forget plot]
table {%
1 0.879310369491577
1 0.844827592372894
};
\addplot [black, forget plot]
table {%
1 0.944444417953491
1 0.944444417953491
};
\addplot [black, forget plot]
table {%
0.875 0.844827592372894
1.125 0.844827592372894
};
\addplot [black, forget plot]
table {%
0.875 0.944444417953491
1.125 0.944444417953491
};
\addplot [black, forget plot]
table {%
2 0.880794256925583
2 0.852459013462067
};
\addplot [black, forget plot]
table {%
2 0.944444417953491
2 0.944444417953491
};
\addplot [black, forget plot]
table {%
1.875 0.852459013462067
2.125 0.852459013462067
};
\addplot [black, forget plot]
table {%
1.875 0.944444417953491
2.125 0.944444417953491
};
\addplot [black, forget plot]
table {%
3 0.880794256925583
3 0.844827592372894
};
\addplot [black, forget plot]
table {%
3 0.944444417953491
3 0.944444417953491
};
\addplot [black, forget plot]
table {%
2.875 0.844827592372894
3.125 0.844827592372894
};
\addplot [black, forget plot]
table {%
2.875 0.944444417953491
3.125 0.944444417953491
};
\addplot [black, forget plot]
table {%
4 0.879310369491577
4 0.852459013462067
};
\addplot [black, forget plot]
table {%
4 0.944444417953491
4 0.944444417953491
};
\addplot [black, forget plot]
table {%
3.875 0.852459013462067
4.125 0.852459013462067
};
\addplot [black, forget plot]
table {%
3.875 0.944444417953491
4.125 0.944444417953491
};
\path [draw=color1, fill=color1]
(axis cs:0.75,0.879310369491577)
--(axis cs:1.25,0.879310369491577)
--(axis cs:1.25,0.944444417953491)
--(axis cs:0.75,0.944444417953491)
--(axis cs:0.75,0.879310369491577)
--cycle;
\path [draw=color2, fill=color2]
(axis cs:1.75,0.880794256925583)
--(axis cs:2.25,0.880794256925583)
--(axis cs:2.25,0.944444417953491)
--(axis cs:1.75,0.944444417953491)
--(axis cs:1.75,0.880794256925583)
--cycle;
\path [draw=color3, fill=color3]
(axis cs:2.75,0.880794256925583)
--(axis cs:3.25,0.880794256925583)
--(axis cs:3.25,0.944444417953491)
--(axis cs:2.75,0.944444417953491)
--(axis cs:2.75,0.880794256925583)
--cycle;
\path [draw=color4, fill=color4]
(axis cs:3.75,0.879310369491577)
--(axis cs:4.25,0.879310369491577)
--(axis cs:4.25,0.944444417953491)
--(axis cs:3.75,0.944444417953491)
--(axis cs:3.75,0.879310369491577)
--cycle;
\addplot [white, forget plot]
table {%
0.75 0.896551728248596
1.25 0.896551728248596
};
\addplot [white, forget plot]
table {%
1.75 0.901639342308044
2.25 0.901639342308044
};
\addplot [white, forget plot]
table {%
2.75 0.896551728248596
3.25 0.896551728248596
};
\addplot [white, forget plot]
table {%
3.75 0.896551728248596
4.25 0.896551728248596
};
\end{axis}

\end{tikzpicture}}
        \caption{\scriptsize{Federated learning (Real)}}
      \label{fig:alexnet_real_acc_fused}
    \end{subfigure}
    \caption{Accuracy of the different models obtained through centralized learning with aggregated data or federated learning with fused local models based on AlexNet. These results are trained ($^t$) and validated ($^v$) with respective simulation datasets ($\mathcal{S}_i^t$, $\mathcal{S}_i^v$) and real datasets ($\mathcal{R}_i^t$, $\mathcal{R}_i^v$) independently.}
    \label{fig:alexnet_accuracy_results}
\end{figure*}

 
\begin{table*}[t]
    \centering
    \caption{Area under ROC curve (AUC) values for the aggregated centralized learning and federated learning approaches.}
    \label{tab:auc_values}
    \small
    \begin{tabular}{@{}cclccccccccc@{}}
        \toprule \\[-.85em]
        \hspace{.23em} & \hspace{.23em} & & \multicolumn{9}{c}{\normalsize{Training datasets (AlexNet, ResNet18)}} \\[+.25em]
        \cmidrule{4-12} \\[-.85em]
        & & & \multicolumn{4}{c}{Centralized learning with aggregated data} &  & \multicolumn{4}{c}{Federated learning} \\[+.25em]
        \cmidrule{4-7} \cmidrule{9-12} \\[-.55em]
        \multicolumn{1}{c}{\multirow{12}{*}{\rotatebox[origin=l]{90}{\hspace{1em} 
        \normalsize{Validation datasets} \hspace{1em}}}} &  &  & $\mathcal{S}_{0,1}^t$ & $\mathcal{S}_{0,2}^t$ & $\mathcal{S}_{1,2}^t$ & $\mathcal{S}_{1,2,3}^t$ & & $\mathcal{S}_{0,1}^t$ & $\mathcal{S}_{0,2}^t$ & $\mathcal{S}_{1,2}^t$ & $\mathcal{S}_{1,2,3}^t$ \\[+0.6em]
        & \parbox[t]{1mm}{\multirow{3}{*}{\rotatebox[origin=c]{90}{Sim}}}
        & \textbf{$\mathcal{S}_{0}^v$} & (0.33, 0.53)  & (0.50, 0.58) & (0.71, 0.56) & (0.52, 0.31) & & (\textbf{0.85}, 0.74) & (\textbf{0.85}, 0.78) & (\textbf{0.85}, \textbf{0.79}) & (\textbf{0.85}, 0.78) \\
        & & \textbf{$\mathcal{S}_{1}^v$}  & (0.33, 0.54) & (0.50, 0.40) & (0.75, \textbf{0.96}) & (0.60, 0.21) & &  (0.94, 0.88) & (0.93, 0.89) & (0.93, 0.86) & (\textbf{0.95}, 0.89) \\
        & & \textbf{$\mathcal{S}_{2}^v$}  & (0.42, 0.64) & (0.50, 0.50) & (0.23, 0.26) & (0.62, 0.23) & & (\textbf{0.96}, 0.84) & (0.95, 0.92) & (0.95, \textbf{0.93}) & (0.95, 0.91) \\
        \\[-0.4em]
        
        & & & $\mathcal{R}_{0,1}^t$ & $\mathcal{R}_{0,2}^t$ & $\mathcal{R}_{1,2}^t$ & $\mathcal{R}_{1,2,3}^t$ & & $\mathcal{R}_{0,1}^t$ & $\mathcal{R}_{0,2}^t$ & $\mathcal{R}_{1,2}^t$ & $\mathcal{R}_{1,2,3}^t$ \\[+0.6em]
        & \parbox[t]{1mm}{\multirow{3}{*}{\rotatebox[origin=c]{90}{Real}}}
        & $\mathcal{R}_0^v$  & (0.75, 0.42) & (0.63,  0.50) & (0.08, 0.33) & (0.58, \textbf{1.00}) & & (\textbf{0.88}, 0.86) & (\textbf{0.88}, 0.75) & (\textbf{0.88}, 0.94) & (\textbf{0.88}, \textbf{1.00}) \\
        & & $\mathcal{R}_1^v$  & (0.35, 0.68) & (0.25, 0.69) & (0.50, 0.48) & (0.70, 0.71) & & (0.83, 0.80) & (\textbf{0.85}, \textbf{0.86}) & (\textbf{0.85}, 0.76) & (\textbf{0.85}, 0.82) \\
        & & $\mathcal{R}_2^v$  & (0.46, 0.45) & (0.51, 0.42) & (0.48, 0.47) & (0.56, 0.67) & & (0.90, 0.70) & (\textbf{0.92}, \textbf{0.87}) & (0.90, 0.73) & (\textbf{0.92}, 0.82) \\
        \bottomrule
    \end{tabular}
\end{table*}


\section{Experimental Results}

This section presents experimental results obtained using data from both simulated and real robots. We first demonstrate the performance of various knowledge-sharing approaches for obstacle avoidance using AlexNet and ResNet18 models before delving into the possibility of sim-to-real knowledge transferability. Finally, we demonstrated the FL method's ability to facilitate lifelong learning by having Husky performing simulation and real-world navigation tasks.

\subsection{Centralized training vs. federated learning}

Our experiments begin by analyzing the performance improvements of federated learning over traditional centralized training with data aggregation. Both simulated and real-world data are evaluated separately in this part. 
Extending the previous works in~\cite{yu2022federated}, by including the AlextNet and ResNet18 in this study, we can compare how the performance will vary for centralized training and FL using different DL models. To accomplish this, we used the data we collected in the simulated hospital ($\mathcal{S}_{0}^t$), office ($\mathcal{S}_{1}^t$), and warehouse ($\mathcal{S}_{2}^t$) to train our model on each dataset and all possible combinations ($\mathcal{S}_{0,1}^t$, $\mathcal{S}_{0,2}^t$, $\mathcal{S}_{1,2}^t$, $\mathcal{S}_{0,1,2}^t$) of two or three of these datasets and validate the models on $\mathcal{S}_{{i,\:i\in\{0,1,2\}}}^v$. Equivalently for the federated learning approach, we run different training rounds in which we simulate that a different subset of robots is collaboratively learning without sharing any actual raw data. Only the models are fused in this approach, and a global model is updated iteratively. In the case of real-world data, we repeat the procedure above with $\mathcal{R}_{0}^t$, $\mathcal{R}_{1}^t$. $\mathcal{R}_{2}^t$ representing the training datasets from rooms, office and laboratory and $\mathcal{R}_{0,1}^t$, $\mathcal{R}_{0,2}^t$, $\mathcal{R}_{1,2}^t$, $\mathcal{R}_{0,1,2}^t$ representing the combinations of two or three of the previous sets while $\mathcal{R}_{{i,\:i\in\{0,1,2\}}}^v$ denotes the corresponding validation data.
\Cref{fig:alexnet_accuracy_results} and \cref{fig:resnet_accuracy_results} report the accuracy of the different models (AlexNet and ResNet18) for centralized learning and FL, respectively. From the results, we found that FL based approach is robust both in a sim and real separately, and its accuracy is competitive with traditional centralized data aggregation training methods for performing vision-based obstacle avoidance tasks.

Along with accuracy, we also calculate the area under the ROC curve (AUC) for each of the scenarios where training is conducted through either the centralized or federated approaches. The results are summarized in~\cref{tab:auc_values} where the AUC values for AlexNet-based and ResNet18-based models are enclosed with a bracket in order. This metric enables a better understanding of the models' reliability. In the context of robotic navigation, there is indeed a cost differential between false negatives over false positives in terms of the robots' integrity. However, from the point of view of performance, false positives can significantly degrade the navigation speed and time, while low-frequency collisions can be avoided with other sensors, considering as well that a false negative is not necessarily consistent over time and multiple observations of the same obstacle are processed before a collision may happen.

For both AlexNet and ResNet18, the AUC results together with the accuracy boxplots show that there is a performance boost with the federated learning approach in contrast to the centralized learning method where all data is first aggregated in a single training set.

\begin{figure}[t]
    \centering
    \setlength{\figurewidth}{.95\linewidth}
    \setlength{\figureheight}{.65\linewidth}
    \scriptsize{
\begin{tikzpicture}

\definecolor{color0}{rgb}{1,0.498039215686275,0.0549019607843137}
\definecolor{color1}{rgb}{0.83921568627451,0.152941176470588,0.156862745098039}
\definecolor{color2}{rgb}{0.12156862745098,0.466666666666667,0.705882352941177}
\definecolor{color3}{rgb}{0.580392156862745,0.403921568627451,0.741176470588235}
\definecolor{color4}{rgb}{0.890196078431372,0.466666666666667,0.76078431372549}

\begin{axis}[
    height=\figureheight,
    width=\figurewidth,
    axis line style={white},
    legend columns = 2,
    legend style = {
        draw opacity=0.2,
        fill opacity=1,
        text opacity=1,
        at={(0.5, 1.2)}, 
        anchor=north, 
        inner sep=2pt, 
        style={column sep=0.1cm}},       
        legend cell align=left,
        ticklabel style={
            /pgf/number format/.cd,
            fixed,
            use comma,
        },
    tick align=outside,
    tick pos=left,
    x grid style={white},
    xmin=-0.15, xmax=3.15,
    xtick style={color=black},
    xtick={0,1,2,3},
    xticklabels={$\mathcal{S}^t_{0,1}$,$\mathcal{S}^t_{0,2}$,$\mathcal{S}^t_{1,2}$,$\mathcal{S}^t_{0,1,2}$},
    y grid style={white!69.0196078431373!black},
    ylabel={Accuracy on $\mathcal{R}^{*,v}$},
    ymajorgrids,
    ymin=0.5, ymax=0.9,
    ytick style={color=black},
    ]
\addplot [only marks, color0, mark=*, mark size=2.5, mark options={solid}]
table {%
0 0.534090936183929
1 0.545454561710358
2 0.541666686534882
3 0.840909063816071
};
\addlegendentry{Centralized Learning AlexNet}
\addplot [only marks, color1, mark=triangle*, mark size=2.5, mark options={solid}]
table {%
0 0.791666686534882
1 0.799242436885834
2 0.784090936183929
3 0.80303031206131
};
\addlegendentry{FL AlexNet}
\addplot [only marks, color2, mark=square*, mark size=2.5, mark options={solid}]
table {%
0 0.621212124824524
1 0.628787875175476
2 0.613636374473572
3 0.651515126228333
};
\addlegendentry{Centralized Learning ResNet}
\addplot [only marks, color3, mark=pentagon*, mark size=2.5, mark options={solid}]
table {%
0 0.64393937587738
1 0.738636374473572
2 0.715909063816071
3 0.742424249649048
};
\addlegendentry{FL ResNet}
\end{axis}

\end{tikzpicture}}
    \caption{Evaluation of sim-to-real capability of the trained models.}
    \label{fig:sim2real_val}
\end{figure}
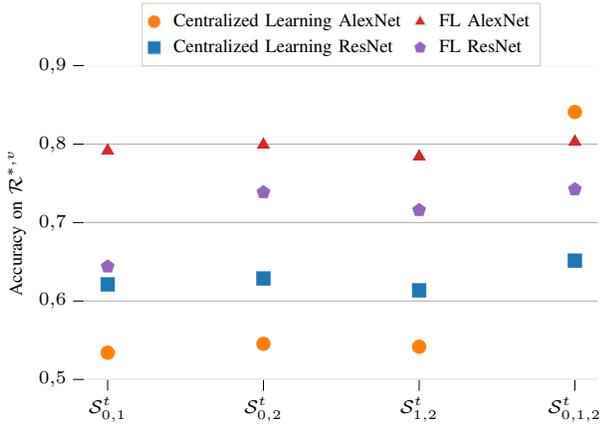

\begin{figure}[t]
     \centering
     \begin{subfigure}[b]{0.48\linewidth}
         \centering
         \includegraphics[width=0.9\textwidth]{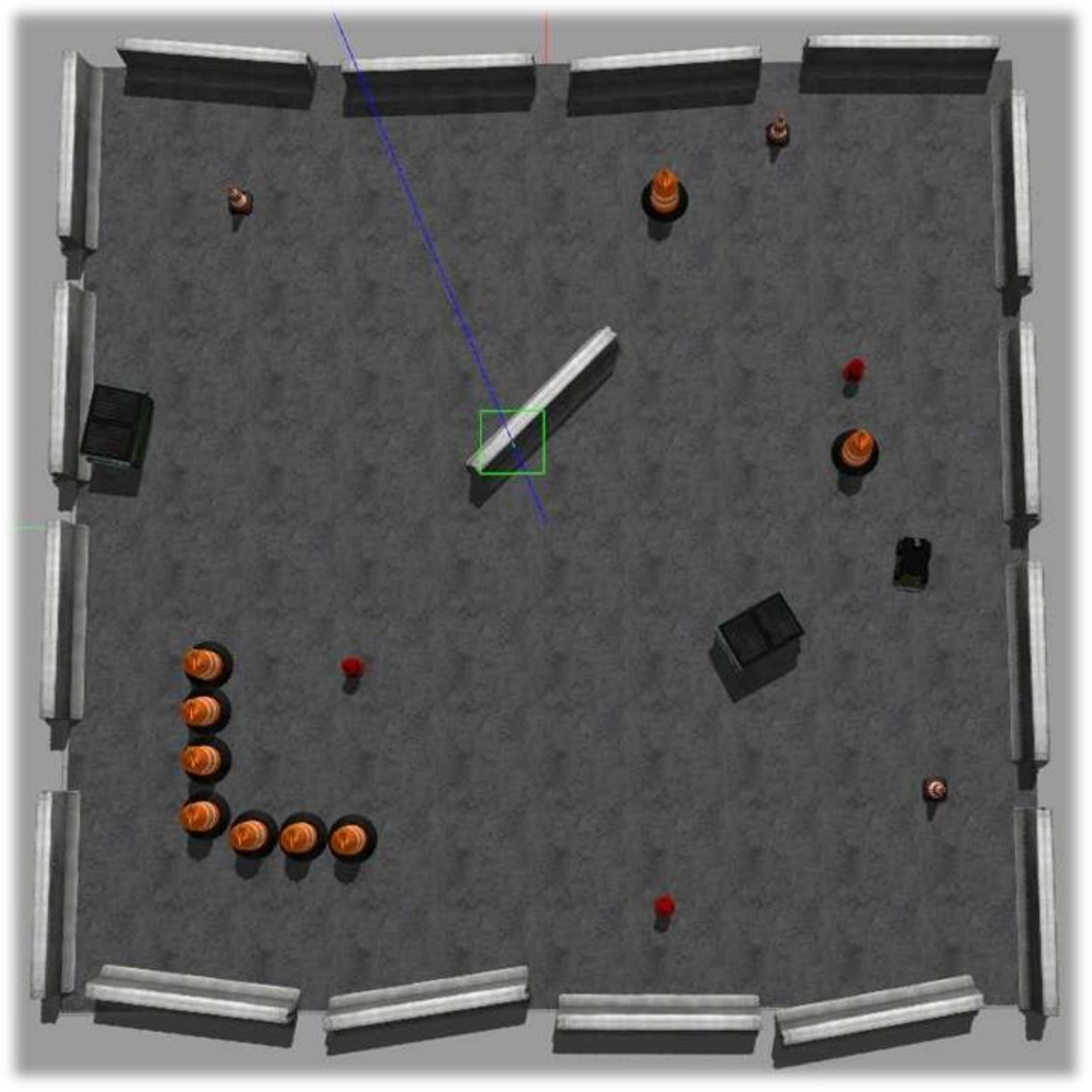}
         \caption{Sim: operating environment}
         \label{fig:husky_sim_env}
     \end{subfigure}
     \hfill
    \begin{subfigure}[b]{0.48\linewidth}
         \centering
         \includegraphics[width=0.9\textwidth]{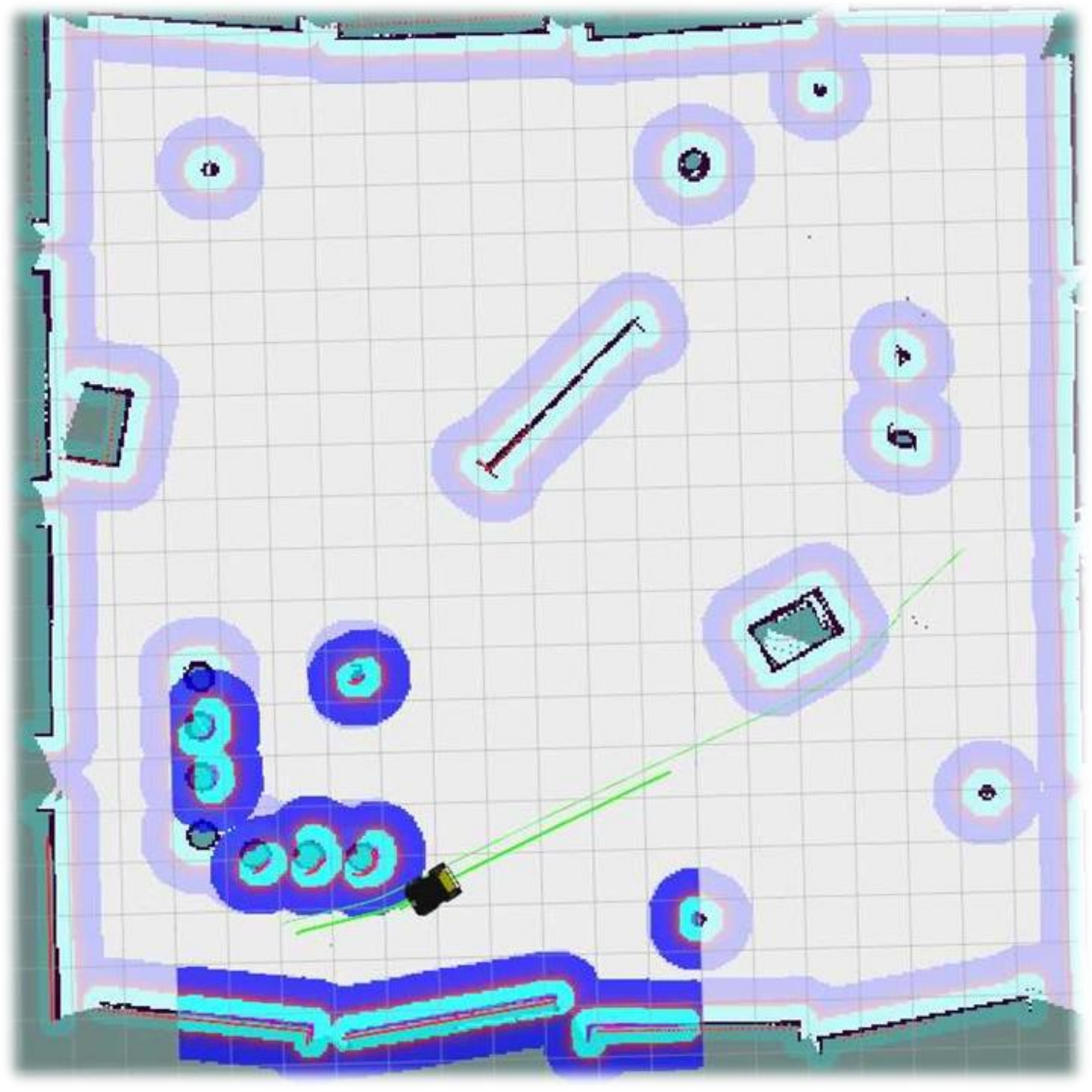}
         \caption{Sim: navigation map}
         \label{fig:husky_sim_map}
     \end{subfigure}
     \vspace{1em}
     \begin{subfigure}[b]{0.48\linewidth}
         \centering
         \includegraphics[width=0.9\textwidth]{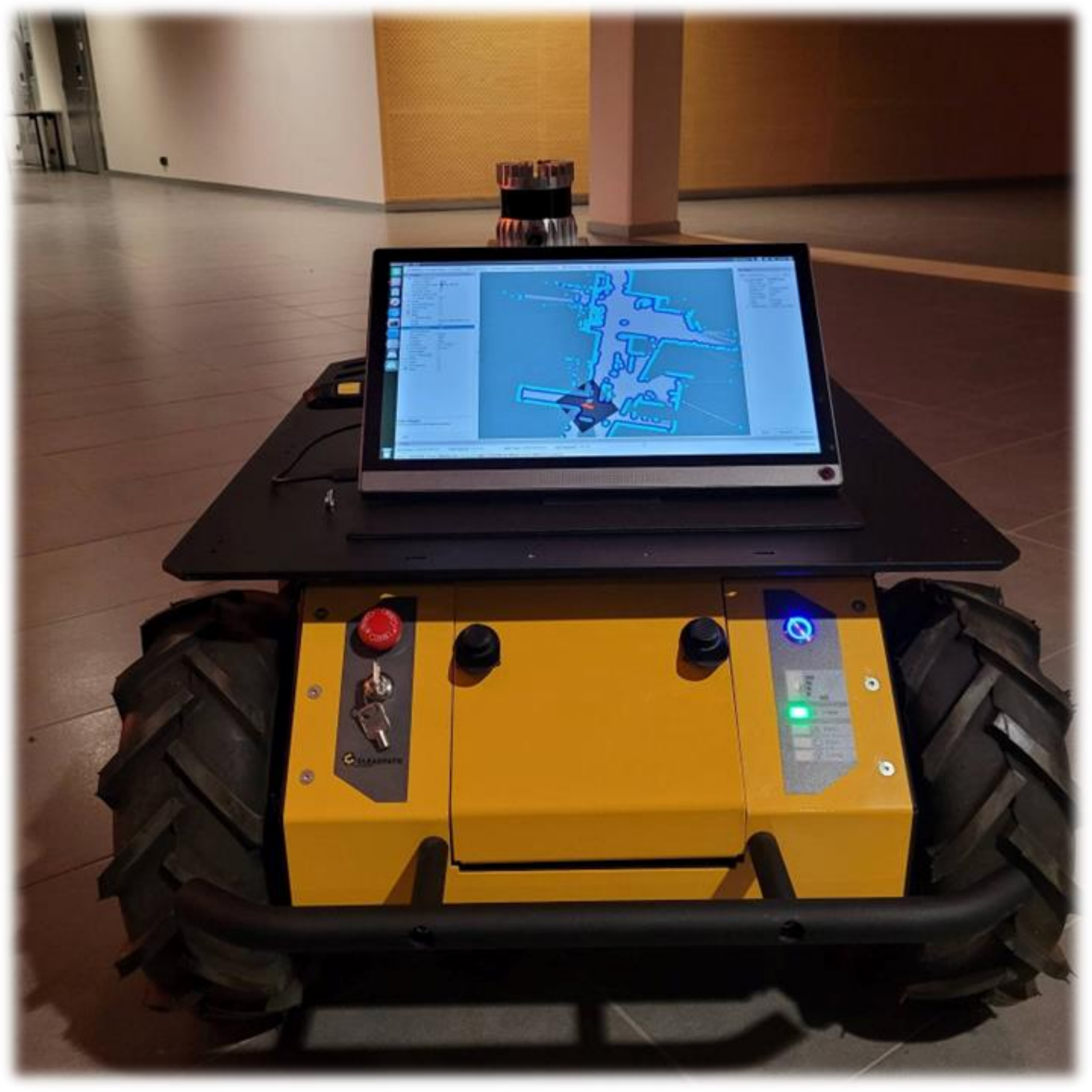}
         \caption{Real: operating environment}
         \label{fig:husky_real_env}
     \end{subfigure}
     \hfill
    \begin{subfigure}[b]{0.48\linewidth}
         \centering
         \includegraphics[width=0.9\textwidth]{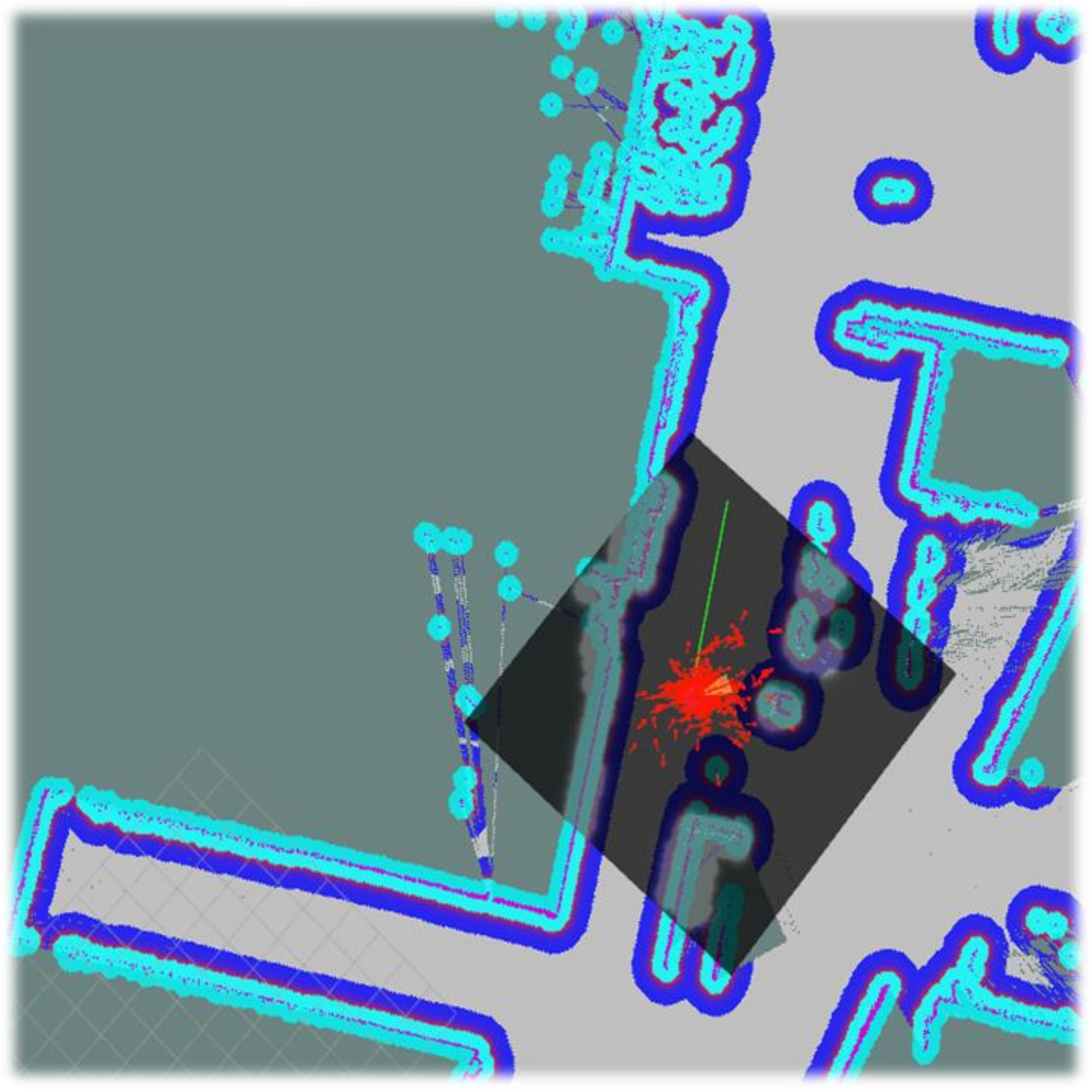}
         \caption{Real: navigation map}
         \label{fig:husky_real_map}
     \end{subfigure}
     \caption{Husky Navigation Environment and Maps}
     \label{fig:husky_nav}
\end{figure}

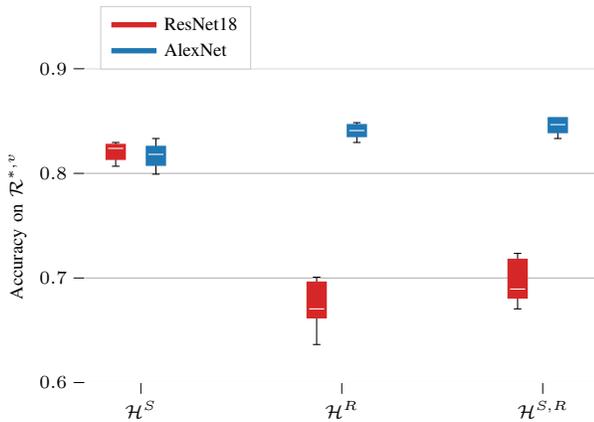
\begin{figure}[t]
    \centering
    \setlength{\figurewidth}{.95\linewidth}
    \setlength{\figureheight}{.65\linewidth}
    \scriptsize{
\begin{tikzpicture}

\definecolor{color0}{rgb}{1,0.498039215686275,0.0549019607843137}
\definecolor{color1}{rgb}{0.83921568627451,0.152941176470588,0.156862745098039}
\definecolor{color2}{rgb}{0.12156862745098,0.466666666666667,0.705882352941177}
\definecolor{color3}{rgb}{0.580392156862745,0.403921568627451,0.741176470588235}
\definecolor{color4}{rgb}{0.890196078431372,0.466666666666667,0.76078431372549}

\begin{axis}[
    height=\figureheight,
    width=\figurewidth,
    legend cell align={left},
    legend style={
      fill opacity=0.8,
      draw opacity=1,
      text opacity=1,
      at={(0.03, 1.2)},
      anchor=north west,
      draw=white!80!black
    },
    axis line style = {white},
    tick align=outside,
    tick pos=left,
    x grid style={white!69.0196078431373!black},
    xmin=-3.075, xmax=48.075,
    xtick style={color=black},
    xtick={2.5,22.5,42.5},
    xticklabels={$\mathcal{H}^{S}$, $\mathcal{H}^{R}$, $\mathcal{H}^{S,R}$},
    ymajorgrids,
    y grid style={white!69.0196078431373!black},
    ylabel={Accuracy on $\mathcal{R}^{*,v}$},
    ymin=0.6, ymax=0.9,
    ytick style={color=black}
]
\addplot [black, forget plot]
table {%
0 0.815340906381607
0 0.806818187236786
};
\addplot [black, forget plot]
table {%
0 0.825757563114166
0 0.829545438289642
};
\addplot [black, forget plot]
table {%
-0.375 0.806818187236786
0.375 0.806818187236786
};
\addplot [black, forget plot]
table {%
-0.375 0.829545438289642
0.375 0.829545438289642
};
\addplot [black, forget plot]
table {%
4 0.809659093618393
4 0.799242436885834
};
\addplot [black, forget plot]
table {%
4 0.823863625526428
4 0.833333313465118
};
\addplot [black, forget plot]
table {%
3.625 0.799242436885834
4.375 0.799242436885834
};
\addplot [black, forget plot]
table {%
3.625 0.833333313465118
4.375 0.833333313465118
};
\addplot [black, forget plot]
table {%
20 0.663825780153275
20 0.636363625526428
};
\addplot [black, forget plot]
table {%
20 0.694128781557083
20 0.700757563114166
};
\addplot [black, forget plot]
table {%
19.625 0.636363625526428
20.375 0.636363625526428
};
\addplot [black, forget plot]
table {%
19.625 0.700757563114166
20.375 0.700757563114166
};
\addplot [black, forget plot]
table {%
24 0.837121188640594
24 0.829545438289642
};
\addplot [black, forget plot]
table {%
24 0.844696998596191
24 0.848484873771667
};
\addplot [black, forget plot]
table {%
23.625 0.829545438289642
24.375 0.829545438289642
};
\addplot [black, forget plot]
table {%
23.625 0.848484873771667
24.375 0.848484873771667
};
\addplot [black, forget plot]
table {%
40 0.682765156030655
40 0.670454561710358
};
\addplot [black, forget plot]
table {%
40 0.715909108519554
40 0.723484873771667
};
\addplot [black, forget plot]
table {%
39.625 0.670454561710358
40.375 0.670454561710358
};
\addplot [black, forget plot]
table {%
39.625 0.723484873771667
40.375 0.723484873771667
};
\addplot [black, forget plot]
table {%
44 0.840909063816071
44 0.833333313465118
};
\addplot [black, forget plot]
table {%
44 0.851325780153275
44 0.852272748947144
};
\addplot [black, forget plot]
table {%
43.625 0.833333313465118
44.375 0.833333313465118
};
\addplot [black, forget plot]
table {%
43.625 0.852272748947144
44.375 0.852272748947144
};
\path [draw=color1, fill=color1, opacity=1, line width=2pt]
(axis cs:-0.75,0.815340906381607)
--(axis cs:0.75,0.815340906381607)
--(axis cs:0.75,0.825757563114166)
--(axis cs:-0.75,0.825757563114166)
--(axis cs:-0.75,0.815340906381607)
--cycle;
\path [draw=color2, fill=color2, opacity=1.0, line width=2pt]
(axis cs:3.25,0.809659093618393)
--(axis cs:4.75,0.809659093618393)
--(axis cs:4.75,0.823863625526428)
--(axis cs:3.25,0.823863625526428)
--(axis cs:3.25,0.809659093618393)
--cycle;
\path [draw=color1, fill=color1, opacity=1.0, line width=2pt]
(axis cs:19.25,0.663825780153275)
--(axis cs:20.75,0.663825780153275)
--(axis cs:20.75,0.694128781557083)
--(axis cs:19.25,0.694128781557083)
--(axis cs:19.25,0.663825780153275)
--cycle;
\path [draw=color2, fill=color2, opacity=1.0, line width=2pt]
(axis cs:23.25,0.837121188640594)
--(axis cs:24.75,0.837121188640594)
--(axis cs:24.75,0.844696998596191)
--(axis cs:23.25,0.844696998596191)
--(axis cs:23.25,0.837121188640594)
--cycle;
\path [draw=color1, fill=color1, opacity=1.0, line width=2pt]
(axis cs:39.25,0.682765156030655)
--(axis cs:40.75,0.682765156030655)
--(axis cs:40.75,0.715909108519554)
--(axis cs:39.25,0.715909108519554)
--(axis cs:39.25,0.682765156030655)
--cycle;
\path [draw=color2, fill=color2, opacity=1.0, line width=2pt]
(axis cs:43.25,0.840909063816071)
--(axis cs:44.75,0.840909063816071)
--(axis cs:44.75,0.851325780153275)
--(axis cs:43.25,0.851325780153275)
--(axis cs:43.25,0.840909063816071)
--cycle;
\addplot [white, forget plot]
table {%
-0.75 0.823863625526428
0.75 0.823863625526428
};
\addplot [white, forget plot]
table {%
3.25 0.818181812763214
4.75 0.818181812763214
};
\addplot [white, forget plot]
table {%
19.25 0.670454561710358
20.75 0.670454561710358
};
\addplot [white, forget plot]
table {%
23.25 0.840909063816071
24.75 0.840909063816071
};
\addplot [white, forget plot]
table {%
39.25 0.689393937587738
40.75 0.689393937587738
};
\addplot [white, forget plot]
table {%
43.25 0.846590936183929
44.75 0.846590936183929
};
\addlegendimage{color=color1, opacity=1.0, line width=2pt}
\addlegendentry{ResNet18}
\addlegendimage{color=color2, opacity=1.0, line width=2pt}
\addlegendentry{AlexNet}
\end{axis}

\end{tikzpicture}}
    \caption{Accuracy of FL fused models based on images collected while Husky was operating navigation tasks in simulated and real-world environments for continual learning purposes}
    \label{fig:nav_acc}
\end{figure}


In terms of sim-to-real performance of the two architectures, we followed the steps in~\cite{yu2022federated} and analyzed the performance of both the centralized and federated learning approaches to an independent real-world dataset. The results showing the potential for vision-based obstacle avoidance inference for both AlexNet and ResNet18 are shown in \cref{fig:sim2real_val}. 
The results indicate that both AlexNet and ResNet18-based obstacle avoidance models implemented with the FL approach can outperform the ones with the centralized data aggregation method in terms of sim-to-real performance. Additionally, AlexNet is more suitable for performing obstacle avoidance tasks in our dataset.

\subsection{Federated continual learning for visual-based obstacle avoidance}

In this section, we describe the results from experiments where the Husky robot has been utilized both in a simulated playpen (see \cref{fig:husky_sim_env}) and in a large-open indoor office environment (see \cref{fig:husky_real_env}) to operate autonomous navigation tasks in their navigation maps (see \cref{fig:husky_sim_map} and in~\cref{fig:husky_real_map}, respectively).

At the same time, while the Husky robot was operating autonomous navigation tasks in the sim and real environments, we collected images continuously for training local models for FL models fusion. By validation on an independent dataset $\mathcal{R}^{*,v}$, we evaluated the global models fused with the local model from simulated husky robot $\mathcal{H}^{S}$ only, the local model from real-world Husky robot $\mathcal{H}^{R}$ only, and the local models of both them $\mathcal{H}^{S, R}$. The results are shown in~\cref{fig:nav_acc}.
Regarding AlexNet, fusing the models either from simulation or real-world can improve the accuracy of the vision-based obstacle avoidance. However, for ResNet18, our results show that fusing the local model from the simulator can improve the performance of the global model more than fusing the one from the real-world Husky. By fusing both of them, the accuracy can be improved, which shows the potential benefit of collaborative learning from both simulated and real robots.


\section{Discussion and Conclusion}
\label{sec:conclusion}

This work presents an FL-based lifelong learning method for vision-based obstacle avoidance among various mobile robots involving both simulated and real-world environments. Rather than applying a single deep neural network, we analyzed the performance of the FL-based method compared with the centralized data aggregation method with two different deep neural networks, providing 
to better generalize results. More specifically, we found that the FL approach can bring competitive accuracy compared to centralized learning across the simulated and real worlds while also delivering inherent benefits in communication optimization and data privacy preservation, enabling collaboration across organizations or users. Additionally, we evaluated the FL method's sim-to-real vision-based obstacle avoidance performance. The result indicates that transferring the obstacle knowledge from simulation to reality using the FL method is more effective and stable. Within the FL-based lifelong learning system, one agent can improve its obstacle avoidance performance by aggregating models from local models in other agents situated either in simulated or real-world environments in our study.

In future work, we will concentrate on utilizing and adapting the FL-based lifelong learning system to perform other robotic navigation tasks with a view toward sim-to-real capabilities. We also find potential in dynamically adjusting the simulation environments based on real-world robot experiences, e.g., adding 3D models of new objects found.


\section*{Acknowledgment}

This research work is supported in part by the Academy of Finland's AutoSOS project (Grant No. 328755) and RoboMesh project (Grant No. 336061).

\bibliographystyle{unsrt}
\bibliography{bibliography}

\end{document}